\documentclass[letterpaper, 10 pt, conference]{ieeeconf}

\IEEEoverridecommandlockouts
\usepackage{amsmath} % assumes amsmath package installed
\usepackage{amssymb}  % assumes amsmath package installed
\usepackage[protrusion=false]{microtype}
\usepackage{tensor}
\usepackage{mathtools}
\usepackage{multicol}
\usepackage{multirow}
\usepackage{subcaption}
\usepackage{makecell}
\usepackage[colorinlistoftodos,prependcaption]{todonotes}
\usepackage{siunitx}
\usepackage{comment}

\usepackage{xcolor} % for textcolor
\usepackage{booktabs}

\makeatletter
\let\NAT@parse\undefined
\makeatother
\usepackage{hyperref}
\title{\LARGE \bf
	Multi-modal curb detection and filtering}

\author{Sandipan Das$^{1,2}$, Navid Mahabadi$^{2}$, Saikat Chatterjee$^{1}$, Maurice Fallon$^3$
	\thanks{$^1$ KTH EECS, Sweden. \texttt{\{sandipan,sach\}@kth.se}\newline%
		$^2$ Scania, Sweden. \texttt{\{sandipan.das,navid.mahabadi\}@scania.com}\newline%
		$^3$ Oxford Robotics Institute, UK. \texttt{mfallon@robots.ox.ac.uk}}}

\usepackage[linesnumbered,ruled,vlined]{algorithm2e}

\usepackage{multirow}
\usepackage[normalem]{ulem}
\usepackage{xargs} 

\newcommand{\hide}[1]{}

\newcommand{\bdmath}{\begin{dmath}}
\newcommand{\edmath}{\end{dmath}}
\newcommand{\beq}{\begin{equation}}
\newcommand{\eeq}{\end{equation}}
\newcommand{\bdm}{\begin{displaymath}}
\newcommand{\edm}{\end{displaymath}}
\newcommand{\bea}{\begin{eqnarray}}
\newcommand{\eea}{\end{eqnarray}}
\newcommand{\beal}{\beq \begin{array}{ll}}
\newcommand{\eeal}{\end{array} \eeq}
\newcommand{\beas}{\begin{eqnarray*}}
\newcommand{\eeas}{\end{eqnarray*}}
\newcommand{\ba}{\begin{array}}
\newcommand{\ea}{\end{array}}
\newcommand{\bit}{\begin{itemize}}
\newcommand{\eit}{\end{itemize}}
\newcommand{\ben}{\begin{enumerate}}
\newcommand{\een}{\end{enumerate}}

\newcommand{\Real}{\mathbb{R}}

\newcommand{\calS}{{\cal S}}

\newcommand{\World}{\mathtt{W}}
\newcommand{\Imu}{\mathtt{I}}
\newcommand{\Camera}{\mathtt{C}}
\newcommand{\Lidar}{\mathtt{L}}

\newcommand{\Base}{\mathtt{{B}}}

\SetCommentSty{algo}
\SetKwInput{KwInput}{Input} % Set the Input
\SetKwInput{KwOutput}{Output} % set the Output

\newcommand\cancel{\bgroup\markoverwith{\textcolor{red}{\rule[0.5ex]{2pt}{0.4pt}}}\ULon}

\makeatletter
\let\NAT@parse\undefined
\makeatother
\usepackage{hyperref}

\begin{document}
	
	\maketitle
	\thispagestyle{empty}
	\pagestyle{empty}
	
	\begin{abstract}
		
		Reliable knowledge of road boundaries is critical for autonomous vehicles navigation. We propose a robust curb detection and filtering technique based on the fusion of camera semantics and dense lidar point clouds. The lidar point clouds are collected by fusing multiple lidars for robust feature detection. The camera semantics are based on a modified EfficientNet architecture which is trained with labeled data collected from onboard fisheye cameras. The point clouds are associated with the closest curb segment with $L_2$-norm analysis after projecting into the image space with the fisheye model projection. Next, the selected points are clustered using unsupervised density-based spatial clustering to detect different curb regions. As new curb points are detected in consecutive frames they are associated with the existing curb clusters using temporal reachability constraints. If no reachability constraints are found a new curb cluster is formed from these new points. This ensures we can detect multiple curbs present in road segments consisting of multiple lanes if they are in the sensors' field of view. Finally, Delaunay filtering is applied for outlier removal and its performance is compared to traditional RANSAC-based filtering. An objective evaluation of the proposed solution is done using a high-definition map containing ground truth curb points obtained from a commercial map supplier. The proposed system has proven capable of detecting curbs of any orientation in complex urban road scenarios comprising straight roads, curved roads, and intersections with traffic isles.
		
	\end{abstract}
	
	% Keep Consistent Tenses:
	
	%in background/lit review: past
	%for experiments carried out: past
	%for results being evaluated currenty: present
	
%\mfallon{Suggested title: }

	\section{Introduction}
	\label{sec:introduction}
	% Write some background about this: Why its important for autonomous vehicles.
	% Write some background about this: Why its important for autonomous vehicles.
Robust environmental perception is a fundamental aspect of autonomous driving that is important for road safety and efficiency and contributes to technical problems such as path planning, control, and localization.
Highly dynamic driving environments can pose critical safety challenges for self-driving vehicles. Objects (stationary or moving), as well as road construction, can change the geometry of the road and result in degradation to localization and planning. Curbs define the road boundary and provide useful information for vehicle navigation; as a result, accurately detecting and tracking them is important. 

% a brief summary of existing research comparing different techniques
Over the past few years, there have been numerous methods proposed to detect and extract curb features using either a single sensor or a combination of sensor modalities. Most curb detection systems use lidar and cameras \cite{BarHillel2014}. 
% LiDAR only
Lidar sensors have been frequently used to detect curb features as the curbs are inherently geometric features
\cite{Wijesoma_2004, Wang2019, fernandez2014road, chen2015velodyne}.
% Camera only
Vision based processing techniques have also been proposed including
\cite{oniga2010polynomial, siegemund2010curb, kellner2014road}.
% LiDAR + Camera fusion
Because lidar and vision have different failure modes, sensor fusion has become popular in recent years which exploits the best properties of both the sensors -- camera images for semantics and lidar for depth information\cite{goga2018fusing, deac2019fusion, baek2020curbscan}.

%\mfallon{I dislike listing 10 papers with no specific reference to the papers. It adds nothing to a lit review}

	\begin{figure} [bth!]
		\vspace{-3mm}
		\centering
		\includegraphics[width=1.\columnwidth, trim =-140 0 -60 10,clip]{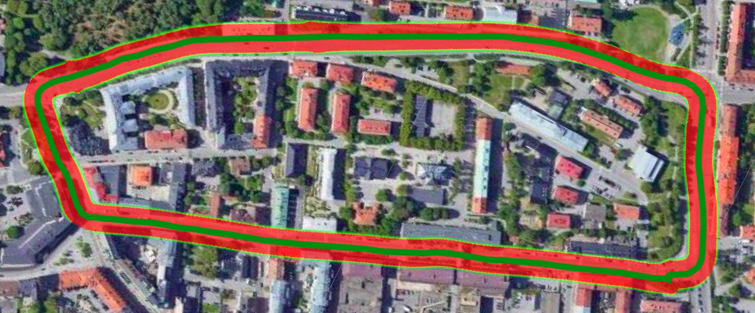}
		\includegraphics[width=1.\columnwidth, trim =5 0 4 0,clip]{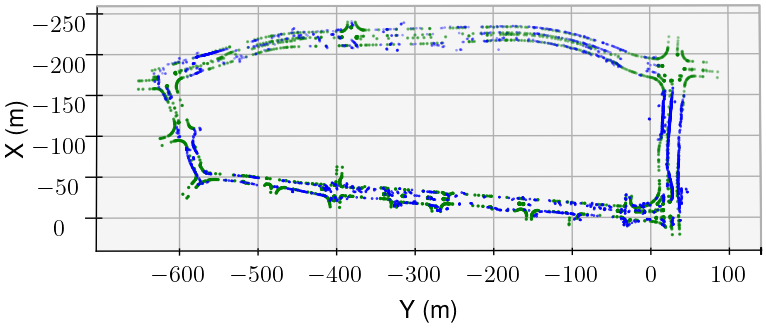}
		\vspace{-5mm}
		\caption{\textit{Top:} The route of the data collection vehicle. \textit{Bottom:} Detected curb features using our proposed method (blue points) and the ground truth curb features (green points) from a commercial map supplier.}
		\label{fig:curb_traj}
		\vspace{-0.5cm}
	\end{figure}
	
	%\mfallon{make the map the same size. add more padding to the top image}

	% Blue color: 48 104 242
	%\mfallon{the boundary of the camera FOV is difficult to understand. Could you make the lines dotted and thicker to make it clearer what they cover?}

	\subsection{Motivation}
	% Why we are doing this: build our own multi-session maps.
	Temporary change to an environment may occur due to various factors. Hence, it is important to incorporate those updates in the mapping module so that the planner and control modules may react accordingly. Consequently, we need an in-house methodology from which we can extract the necessary features of an environment into our mapping module. In addition, we would have the ability to create curb maps in restricted areas to which commercial map suppliers do not have access.

	\subsection{Contribution}
	\label{sec:contribution}
	Our work falls under the category of association of camera semantics with lidar depth. We show the results of our proposed method in \autoref{fig:curb_traj} and to achieve that our specific contributions are:
	\begin{itemize}
		%\item  \mfallon{this is not a contribution its just the type of method ... you cant say `contributions' for everything you do. you say contributions for the NOVEL things you added to the public knowledge base.}
		\item Detection and association of multiple curbs with unsupervised DBSCAN clustering.    
		\item Outlier removal using Delaunay-based filtering method which needs less parametric tuning than RANSAC-based polynomial fitting regression constraints.
	\end{itemize}

	%\mfallon{Stop saying `our map supplier'. This is an academic paper - not a tech report. You should say `a commercial supplier'. similarly you dont speak for Saikat and I when you say `our'.}

	%\mfallon{furthermore, the motivation isnt because your map suppliers maps are bad. the motivation is that prior maps change and need to be updated}

	\section{Related Work}
	\label{sec:related-work}
	
	Aspects of curb detection have been studied extensively in the context of autonomous driving. Multiple solutions have been proposed that use camera, lidar, or fusion of both. 
	%\mfallon{you say camera, lidar and fusion of both in a dozen place in the first page... remove some of them. This is the standard self-driving sensor configuration}
	
	In the early 2000s, traditional computer vision methods were proposed for curb detection. In \cite{aufrere2003multiple}, the authors proposed a histogram filter with a threshold to locate curb points. Whereas, in \cite{Wijesoma_2004}, a Kalman filter based tracking of the threshold-based detected curb points was proposed. By the mid-2000s, Hough transform based method \cite{kim2007outdoor} on 2D lidar data was explored under the assumption that the terrain was flat. Polyfit on a digital elevation map (DEM) created from the stereo vision was explored in  \cite{oniga2010polynomial, siegemund2010curb}, while the authors in \cite{stuckler2008lane} used height images from lidars to create DEM and applied polynomial fitting. In \cite{kellner2014road} the authors proposed to create a DEM from ego-motion and filtering for curb points. As 3D lidar was adopted by researchers after the mid-2000s, a lot of the work for curb detection was based on ground plane segmentation on 3D data \cite{el2011detection, fernandez2014road, chen2015velodyne}.
	
	In recent years, with better network architectures for semantic segmentation, works began to fuse camera semantics with lidar depth \cite{goga2018fusing, deac2019fusion}. Instead of relying on geometric attributes of curbs, the semantic information identified specific curb pixels. Then the depth was obtained from the lidar projected point clouds in the image plane. A recent work, CurbScan \cite{baek2020curbscan}, proposed to fuse an additional ultrasonic sensor for lateral
	distance information with curb tracking using a Kalman filter.
	
	In contrast to prior works, we propose an agglomerative unsupervised clustering to detect multiple curbs in unfiltered point clouds from re-projected camera semantics. We also apply a Delaunay-based filter to remove outliers among the detected curb points. Finally, we transform all the detected curbs into the global frame using GNSS (Global Navigation Satellite System) and check their consistency quantitatively with respect to the ground truth (GT) curb points obtained from a prior hand annotated HD curb map.

	\section{Methodology}
	\label{sec:methodology}
	
	Prior calibration of the sensors is a fundamental prerequisite for sensor fusion. Additionally, an important feature of our fusion technique for feature association is that the lidar point clouds are motion corrected and transformed so as to be equivalent to that recorded at the time stamp of an available camera frame using the approach introduced in \cite{wisth2021unified}. This helps in data from all the sensors need to be properly time synchronized. In the following sub-sections we outline the procedure we adapted for our approach. 

	\subsection{Sensor setup and reference frames}
	\begin{figure}
		\centering
		\vspace{2mm}
		
		\includegraphics[width=0.75\columnwidth]{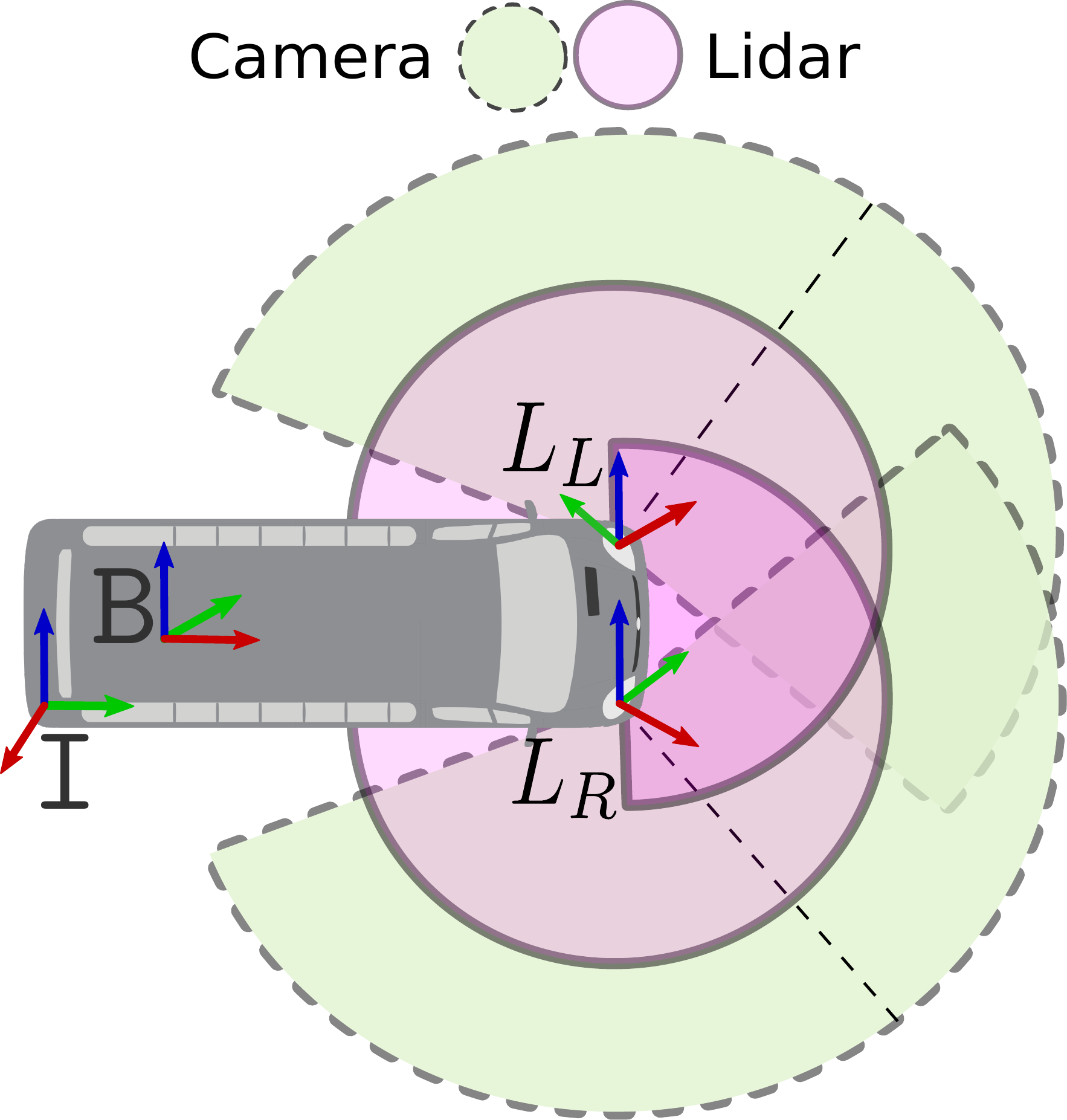}
		%\vspace{-5mm}
		\caption{Illustration of the FoV of two front lidars and four cameras positioned around the data collection vehicle. However, only the two front facing cameras were used for our experiment. The vehicle base frame $\Base$ is located at the center of the rear axle. The sensor frames of the lidars are $\Lidar_{L}$ and $\Lidar_{R}$, representing front-left and front-right lidars respectively. The IMU frame of the GNSS is represented as $\Imu$. }
		\label{fig:scania-bevda}
		\vspace{-0.4cm}
	\end{figure}
	
	The data collection vehicle consisted of two lidars and two cameras. The reference frames and the field of view (FoV) of the sensors are shown in \autoref{fig:scania-bevda}. The vehicle base frame $\Base$ is located on the center of the rear-axle of the vehicle. Sensor readings from lidars and cameras are represented in base frame $\Base$, as $\tensor[_\Base]{\Lidar}{_{k}}$ and $\tensor[_\Base]{\Camera}{_{k}}$ respectively, where $k$ represents the location of the sensor in the vehicle. For example, $\Lidar_{L}$ and $\Lidar_{R}$ represent the front left and front right lidar frames. The pose information from the GNSS is reported in the UTM frame, $\World$ (World frame). The IMU frame, $\Imu$, of the GNSS is shown in \autoref{fig:scania-bevda}. All the inter-sensor transformations are carried out using prior calibration parameters.

	\subsection{Semantic association with lidar points}
	\label{sebsection:semantic-association}
	
	The semantic features from lidar were extracted by associating the projected point clouds with the segmented camera images. The image segmentation was performed using a trained Efficient-Net model.
	%\mfallon{you are using LiDAR and lidar and Lidar in different places. I encourage `lidar' - but please be consistent}
	
	\subsubsection{Efficient-Net for semantic segmentation}
	
	Since AlexNet \cite{alexnet_2012} won the 2012 ImageNet competition, Convolutional Neural Networks (ConvNets) have been increasingly used as the de facto standard for image segmentation tasks. Although higher accuracy is critical for autonomous driving applications, we have already hit the hardware memory limit for the image segmentation tasks using ConvNets. Thus further accuracy gains will require better efficiency. 
	
	The authors of EfficientNet \cite{tan2019efficientnet} illustrated that model scaling can be achieved by carefully balancing network depth, width, and resolution, leading to a better performance with the fixed amount of computation resources. Based on this study, we used a modified architecture of EfficientNet for semantic segmentation. The network architecture backbone is shown in \autoref{tab:efficientNet}, where each row describes a stage $i$ with $\hat{L}_{i}$ layers, with input resolution $\langle\hat{H}_{i}, \hat{W}_{i}\rangle $ and output channels $\hat{C}_{i}$. MBConv layers represent mobile inverted bottlenecks from the MobileNetV2 architecture \cite{sandler2018mobilenetv2}, where squeeze-and-excitation optimization \cite{hu2018squeeze} has also been added on top of it. To upsample the network output to its original input resolution a bilinear interpolation was used in the decoder architecture.

	\begin{table}[!h]
		\vspace{0.3cm}
		\normalsize
		\centering
		\resizebox{\columnwidth}{!}{
			\begin{tabular}{c|cccc} \toprule
				Stage & Operator & Resolution & \#Channels & \#Layers \\
				\( i \) & \( \hat{\mathcal{F}}_{i} \) & \( \hat{H}_{i} \times \hat{W}_{i} \) & \( \hat{C}_{i} \) & \( \hat{L}_{i} \) \\
				\midrule
				1 & Conv3x3 & \( 640 \times 1024 \) & 32 & 1 \\
				2 & MBConv1, k3x3 & \( 320 \times 512 \) & 16 & 1 \\
				3 & MBConv6, k3x3 & \( 160 \times 256 \) & 24 & 2 \\
				4 & MBConv6, k5x5 & \( 80 \times 128 \) & 40 & 2 \\
				5 & MBConv6, k3x3 & \( 40 \times 64 \) & 80 & 3 \\
				6 & MBConv6, k5x5 & \( 40 \times 64 \) & 112 & 3 \\
				7 & MBConv6, k5x5 & \( 20 \times 32 \) & 192 & 4 \\
				8 & MBConv6, k3x3 & \( 20 \times 32 \) & 320 & 1 \\
				9 & Segmentation Head & \( 640 \times 1024 \) & 24 & - \\
				\bottomrule
		\end{tabular}}
		\caption{ EfficientNet-B0 baseline network architecture.}
		\label{tab:efficientNet}
		\vspace{-0.5cm}
	\end{table}
	
	%\cesar{Add the training and testing result metrics. If possible add per class training, test, validation accuracy with IoU. Particularly for the following classes mentioned in Table \ref{tab:efficientNet-results} if available. Discuss!}
	To achieve robustness we trained the EfficientNet model on a diversified set of scenarios (city driving, snowy conditions, dessert area, suburban area). The overall training accuracy is 79.10\% and the mean IoU (Intersection over Union) is 0.495. For the curb class the accuracy is 67.20\% and the IoU is 0.557. The results of the segmentation on a sample frame can be visualized in \autoref{fig:image_semantics}. 
%		\begin{table}[h]
%		\normalsize
%		\centering
%		\resizebox{0.75\columnwidth}{!}{
%			\begin{tabular}{l|cc} \toprule
%				\multirow{2}{*}{\makecell[l]{Class}} & \multicolumn{2}{c}{\makecell[c]{Validation}}\\
%				\cline{2-3}
%				& \textbf{Accuracy (\%)} & \textbf{Mean IoU}\\
%				\midrule
%				Drivable Road & 96.20 & 0.834\\
%				Building & 92.80 & 0.797\\
%				Car & 91.30 & 0.774\\
%				Side Walk & 79.30 & 0.675\\
%				Curb & 67.20 & 0.557\\
%				Fence & 52.30 & 0.379\\
%				\bottomrule
%		\end{tabular}}
%		\caption{Semantic segmentation results using modified EfficientNet-B0 Network with Training accuracy: $\sim$80\% and Mean IoU: $\sim$50\%.}
%		\label{tab:efficientNet-results}
%		\vspace{-3mm}
%	\end{table}
	
	% https://stackoverflow.com/questions/31089265/what-are-the-main-references-to-the-fish-eye-camera-model-in-opencv3-0-0dev
	
	\subsubsection{Association of curb semantics with lidar depth}
\begin{figure}[!h]
		\centering
		\vspace{-0.2cm}
		\begin{multicols}{2}
			\includegraphics[width=1\columnwidth, trim = 0 100 0 0,clip]{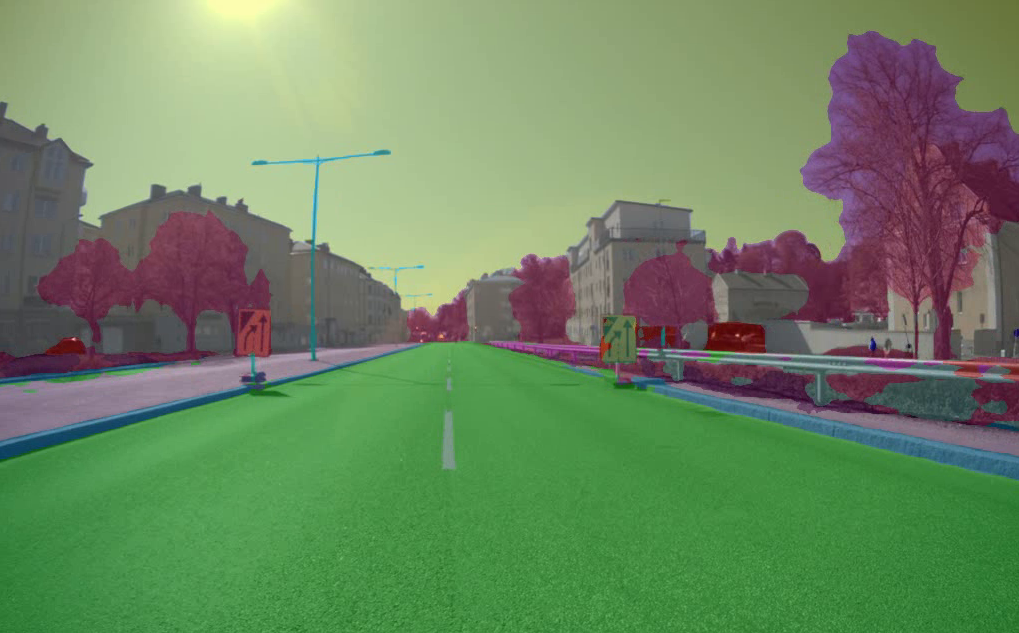}
			\subcaption{Semantic segmentation results using our modified EfficientNet \cite{tan2019efficientnet}.}
			\label{fig:image_semantics}
			\includegraphics[width=1\columnwidth, trim = 0 0 0 50,clip]{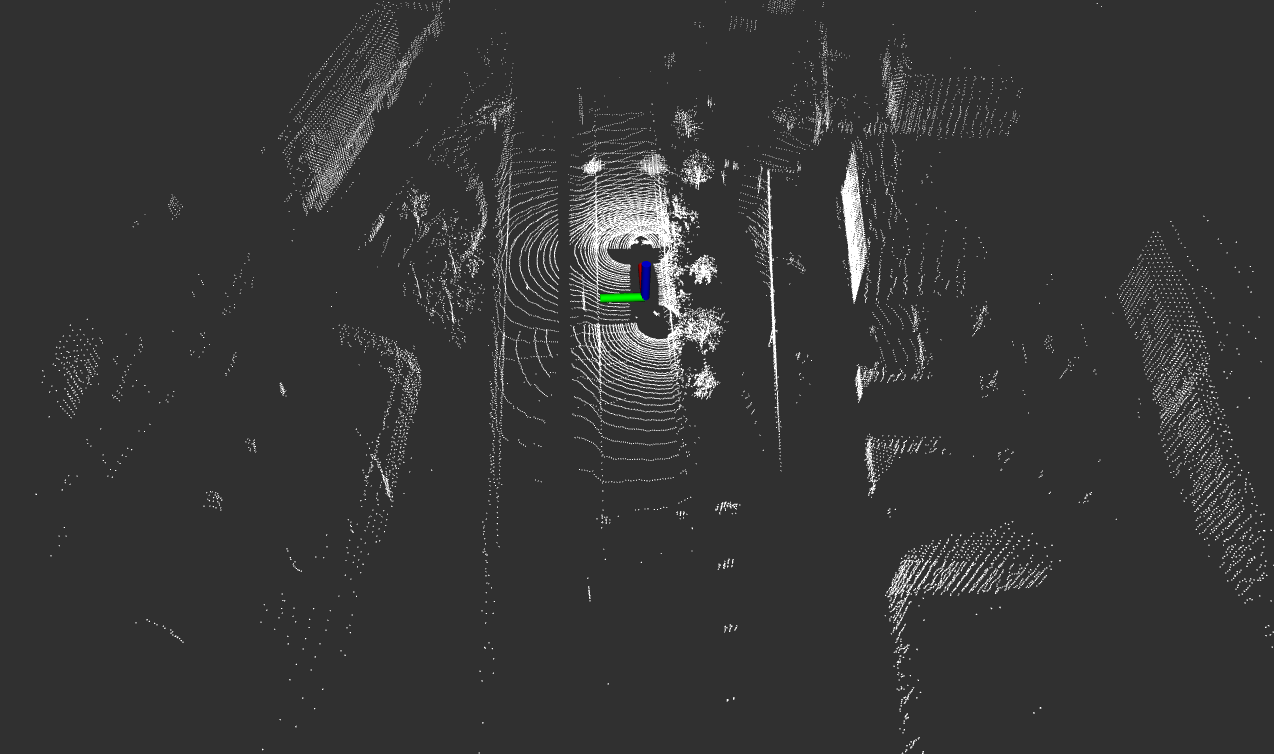}
			\subcaption{Fused lidar point clouds from lidar sensors.}
			\label{fig:lidar_fused}
			\includegraphics[width=1\columnwidth, trim = 0 100 0 0,clip]{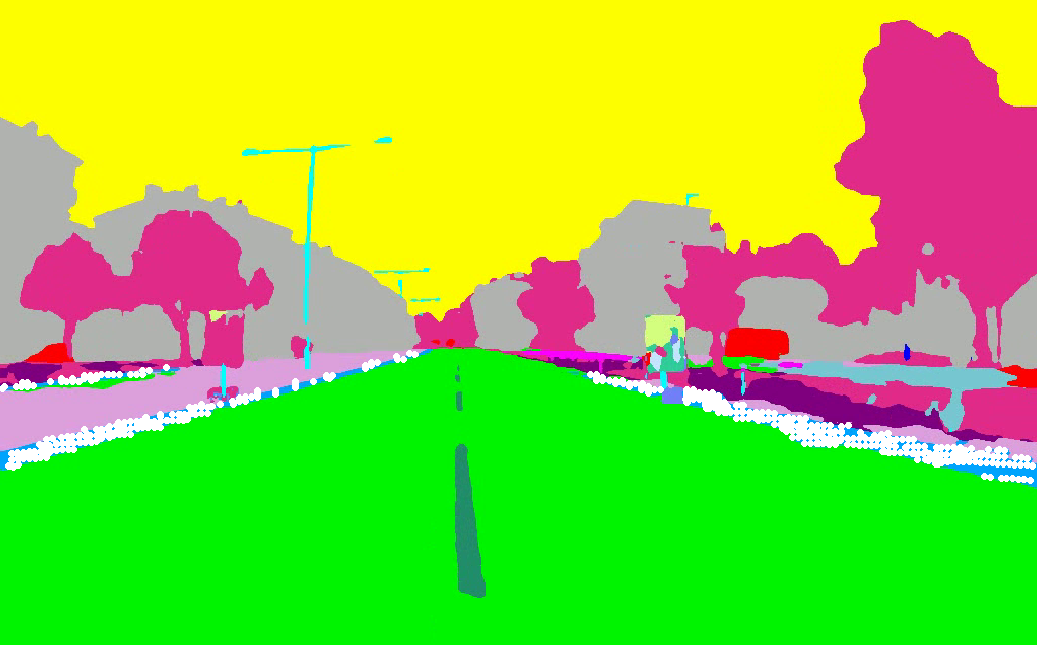}
			\subcaption{Lidar point clouds (white points) overlaid on the segmented curb pixels.}
			\label{fig:lidar_projection}
			\includegraphics[width=1\columnwidth, trim = 0 0 0 50,clip]{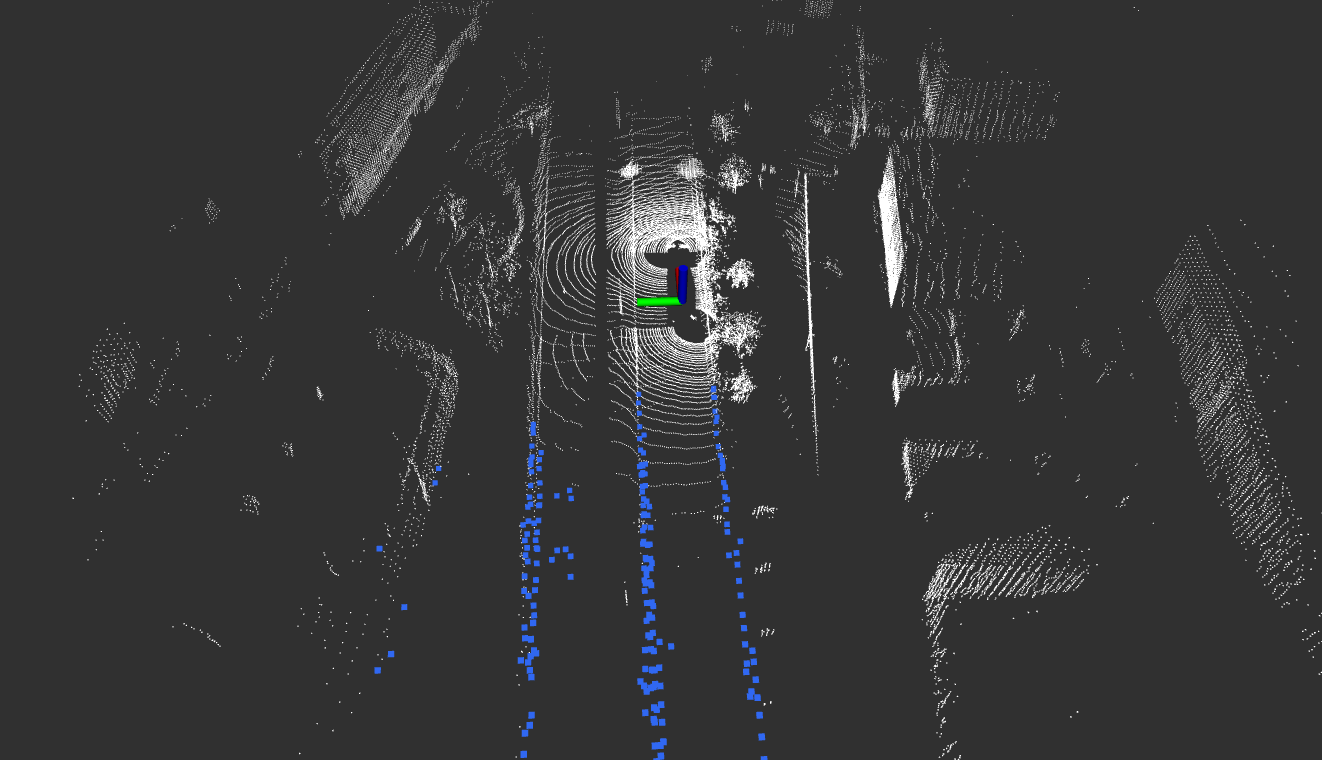}
			\subcaption{Curb semantics (blue points) with the fused point cloud.}
			\label{fig:lidar_reprojection}
		\end{multicols}
		\vspace{-0.4cm}
		\caption{Semantic association with lidar point cloud.}
		\label{fig:lidar-semantic-matching}
		\vspace{-0.5cm}
	\end{figure}

	The cameras mounted on our platform have fisheye lenses. Hence, we extracted the curb points by doing a fish-eye projection \cite{kannala2006generic} of the fused lidar points in image space and chose points closer to the curb pixels within a bound of $\pm$3 pixels. Finally, the curb points are re-projected back to the base frame $\Base$. The results of curb extraction from semantic association can be visualized in \autoref{fig:lidar_reprojection}.
	
	\subsection{Unsupervised clustering and filtering}
	\label{sebsection:clustering}
	The point cloud association technique may give us noisy points due to various reasons such as synchronization quality of the logs, calibration parameters, or camera projection model. To remove outliers we can apply filtering based on the geometric structures of the curbs. However, since we do not know the pre-defined number of curbs in advance, it makes it difficult to apply a polynomial fitting on the extracted curb points. To overcome this problem we first find a set of unsupervised clusters and associate newly detected curb points to the relevant clusters based on spatial density.
	 
	\subsubsection{Iterative cluster association}
\begin{figure}[hbt!]
		\centering
		\vspace{-2mm}
		\includegraphics[width=1\columnwidth]{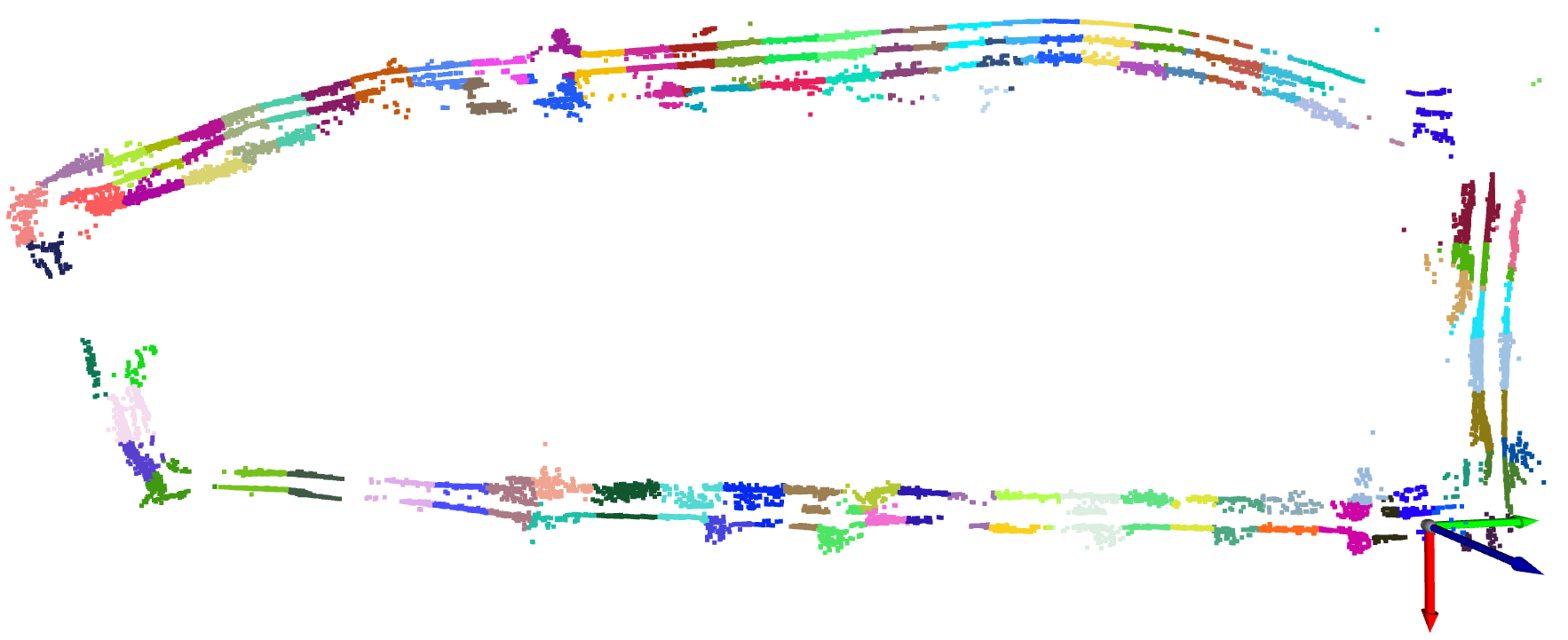} 
		\caption{Iterative feature point clustering using DBSCAN \cite{ester1996density}. Random colors indicate different clusters detected.}
		%\textit{Bottom:} Filtered curb features (blue points) and the ground truth curb features (green points) from our HD maps supplier. The filtering is performed on the clusters after outlier removal using RANSAC. A third order polynomial fitting is also performed on the inlier points based on which noisy points are removed.
		\label{fig:curb-extraction}
	\end{figure}

	We iteratively chose the extracted curb point clouds and applied unsupervised spatial clustering. The boundary points of a cluster are the points which are furthest from the cluster centroids. 
	If the  $L_2$-norm of the boundary points in the new clusters were less than a pre-defined threshold from the old cluster boundary points we merged the clusters. This operation helped in identifying the number of curb segments. The result of our clustering with DBSCAN \cite{ester1996density} is shown in \autoref{fig:curb-extraction}. 
	 
	\subsubsection{Delaunay filtering}
	Let $\calS$ be a set of points in $\Real^n$ with distance Euclidean function $d$. The Voronoi diagram \cite{brandt1992continuous} is the partition of the space into Voronoi regions, $\displaystyle R$, such that 
	\begin{equation}
		{\displaystyle R_{k}=\{s\in \calS\mid d(s,R_{k})\leq d(s,R_{j})\;\forall\;j\neq k\}}.
	\end{equation} 
	Voronoi graphs have been used in motion planning algorithms for obstacle avoidance \cite{vornoi_mp_1989, garber2004constraint}. But in the context of filtering of curb points, this is a novel approach. Delaunay triangulation is the dual of the Voronoi diagram. Delaunay triangulation \cite{fortune1995voronoi} on $\calS$, DT($\calS$) is a triangulation such that no point in $\calS$ is inside the circum-hypersphere of any $n$-simplex in DT($\calS$).
	For $\Real^3$ it is called Delaunay tetrahedralization. The Voronoi vertex corresponding to a Delaunay tetrahedron is the center of the circumscribing sphere of the tetrahedron. Let $[x \quad y \quad z]$ be the center and the four points of the tetrahedron be $[x_i \quad y_i \quad z_i], \forall i=1(1)4$. Then the center is found by solving the following equation:
	\beq
	\begin{aligned}
		&(x-x_1)^2 + (y-y_1)^2 + (z-z_1)^2 = \\ &(x-x_j)^2 + (y-y_j)^2 + (z-z_j)^2  , \forall j=2,3,4\\
		\implies &2(x_j-x_1)x + 2(y_j-y_1)y + 2(z_j-z_1)y = \\&(x_j^2 + y_j^2 + z_j^2) - (x_1^2 + y_1^2 + z_1^2) , \forall j=2,3,4
		\label{eq:dt}
	\end{aligned}
	\eeq
	
	%Only Sweden prefers a lane width of 3.75 m at reference speeds of 90 and 110 km/h. Some countries (the Netherlands, Japan) use traffic lanes of 3.25 m width for motorways in urban areas combined with a lower design speed. Hence the radii for our fitering was set at 5 mt to remove any outliers from a cluster segment.
	
\begin{figure}[h!]
		\centering
		\vspace{0.14cm}
		\begin{multicols}{2}
			\includegraphics[width=1.0\linewidth]{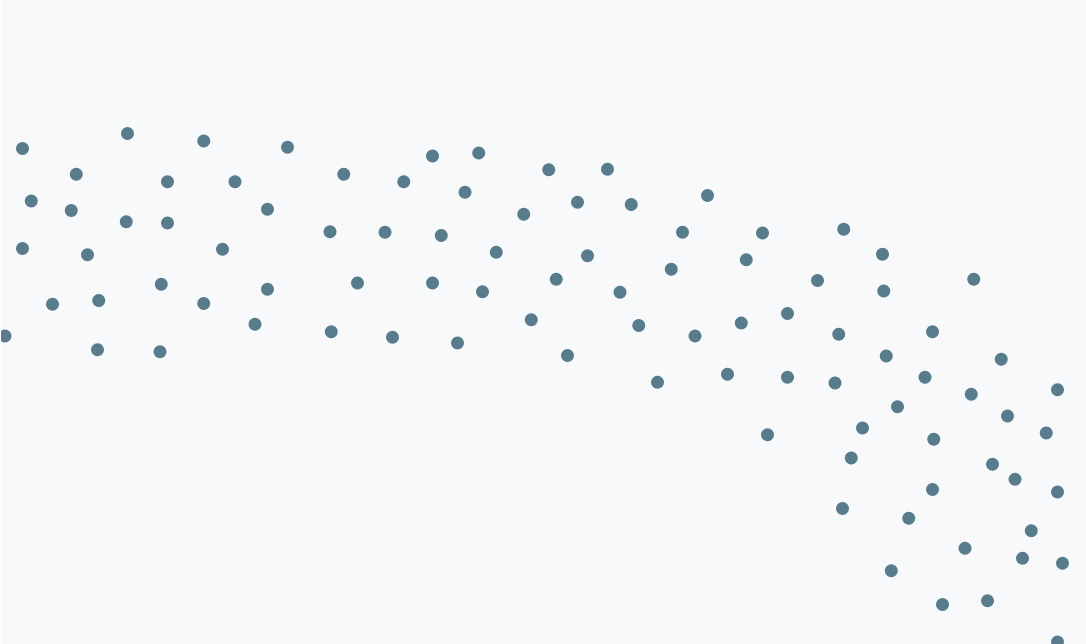}
			\subcaption{Unfiltered curb point cloud}
			\label{fig:curb_raw}
			\includegraphics[width=1.0\linewidth]{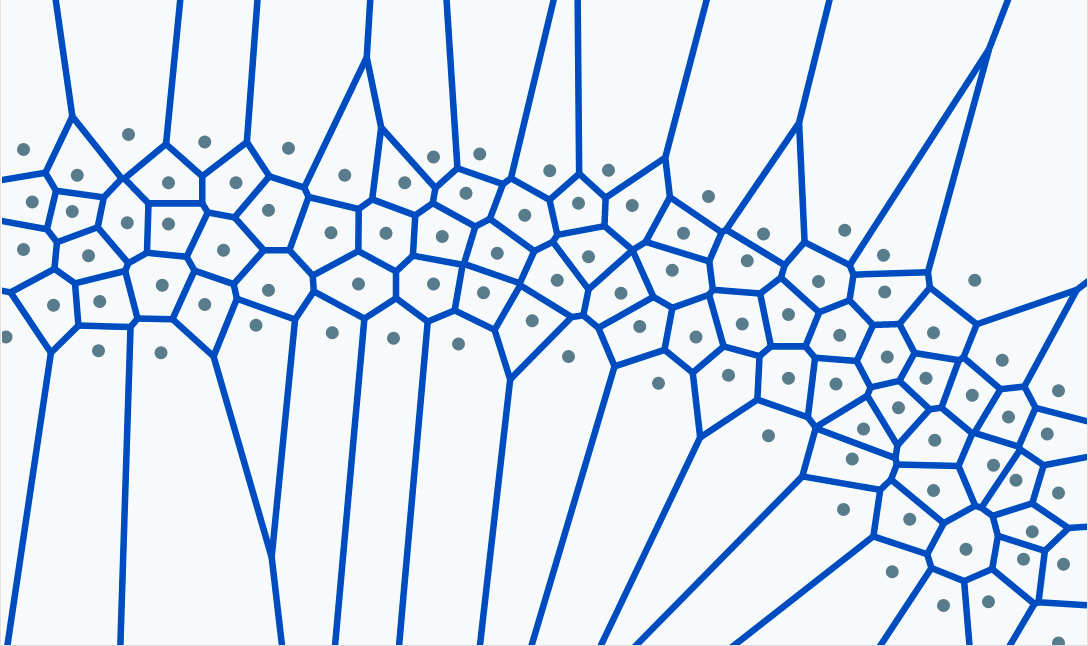}
			\subcaption{Voronoi regions}
			\label{fig:curb_vornoi}
			\includegraphics[width=1.0\linewidth]{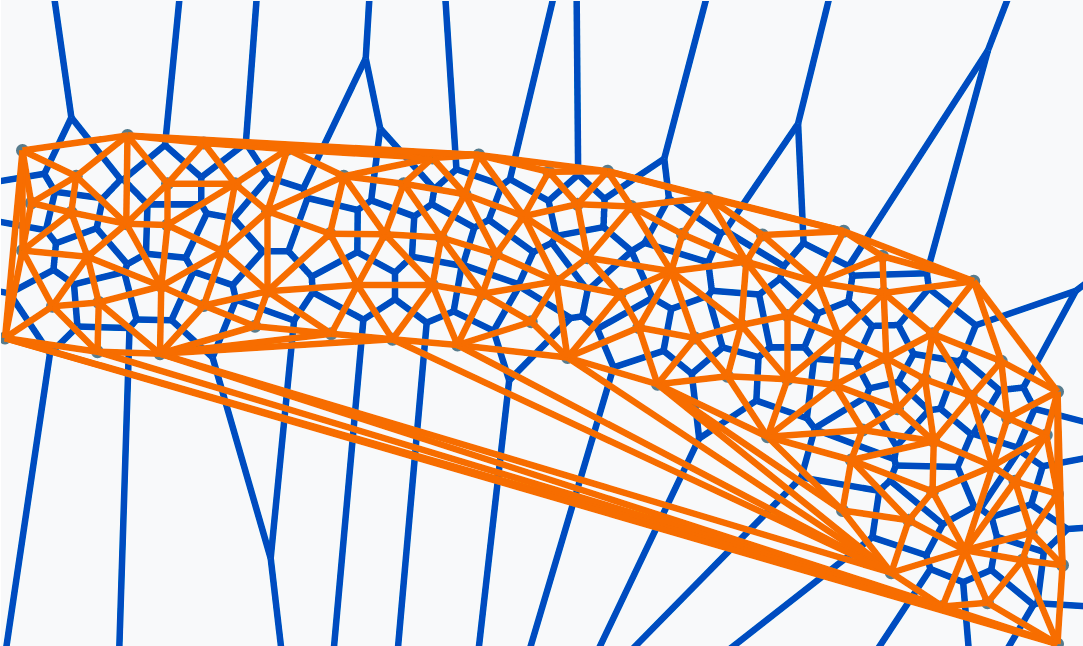}
			\subcaption{Delaunay tetrahedron}
			\label{fig:curb_delaunay}
			\includegraphics[width=1.0\linewidth]{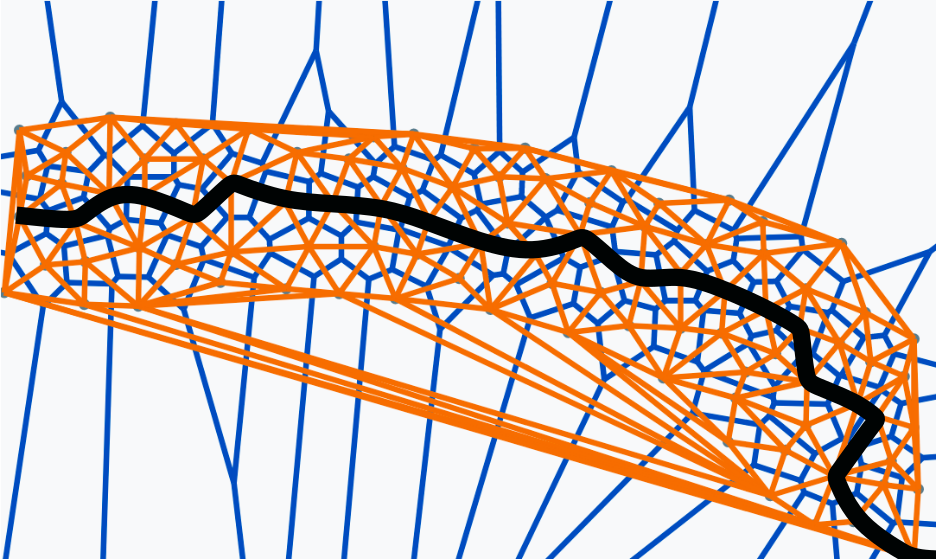}
			\subcaption{Medial axis}
			\label{fig:curb_delaunay_original_data}
		\end{multicols}
		\vspace{-0.6cm}
		\caption{Curb inlier selection with Delaunay filtering.}
		\label{fig:delaunay_filtering}
		\vspace{-0.56cm}
	\end{figure}

	Voronoi sub-graph of the Delaunay tetrahedra is computed by filtering large radii of circumscribing spheres from the centers computed. This removes tetrahedra outside the point volume and removes the outliers. The shortest Euclidean path in the Voronoi sub-graph connecting the start and the end points give us the medial axis. The points closer to the medial axis gives us the filtered point cloud corresponding to the curbs. An illustration of the process is shown in \autoref{fig:delaunay_filtering}.
	%The medial axis is the geometric location of the centers of the maximum sphere inscribed in the tetrahedron.
	
	\subsubsection{RANSAC filtering}
	RANSAC \cite{ransac_1981} is an iterative algorithm for the robust estimation of parameters from a subset of inliers from the complete data set. For fitting the curb points to an estimator we used a third-order polynomial. However, an automatic parameter tuning to estimate the degree of the polynomial was performed for each curb segment to find out the best set of inliers.

	\section{Experimental Results}
	
	\label{sec:results}
	
	\subsection{Dataset}
	
	We collected the data from a Scania autonomous bus platform mounted with two lidars and two front facing cameras for a route length of $\approx$1.5 Kms. The ground truth (GT) curb features are provided by a map supplier. All the sensing data was synchronized using PTP (Precision Time Protocol) synchronization and converted to rosbags for evaluation. We evaluated the generated curb points by manually selecting the corresponding GT curb points. The point cloud selection tool for association was developed using open3d \cite{zhou2018open3d}. Since manual association is a tedious process we also propose a mechanism for automatic evaluation of the clusters. The clustering algorithms are evaluated offline from scikit-learn \cite{scikit-learn} package.
	
%	\begin{figure}[h!]
%		\centering
%		\vspace{-0.3cm}
%		\begin{multicols}{2}
%			%\includegraphics[width=1\columnwidth]{figures/images/curbs_segment_selection.png}
%			%\includegraphics[width=1\columnwidth]{figures/images/curbs_dvsp.png}
%			\includegraphics[width=1.0\linewidth]{figures/images/curbs_selection_gt.png}
%			\subcaption{GT Curb segment selection}
%			\label{fig:curb_selection_gt}
%			\includegraphics[width=1.0\linewidth]{figures/images/curbs_selection_gen.png}
%			\subcaption{Selection in our data}
%			\label{fig:curb_selection_gen}
%		\end{multicols}
%		\vspace{-0.5cm}
%		\caption{Curb segment selection (gray area) in GT and our generated data with our point cloud selection tool.}
%		\label{fig:curb_selection_tool}
%		\vspace{-.69cm}
%	\end{figure}

	\subsection{Manual segment-wise association and evaluation}
\begin{figure}[hbt!]
		\centering
		%\vspace{-0.3cm}
		\includegraphics[width=1\columnwidth, trim = 10 0 0 0,clip]{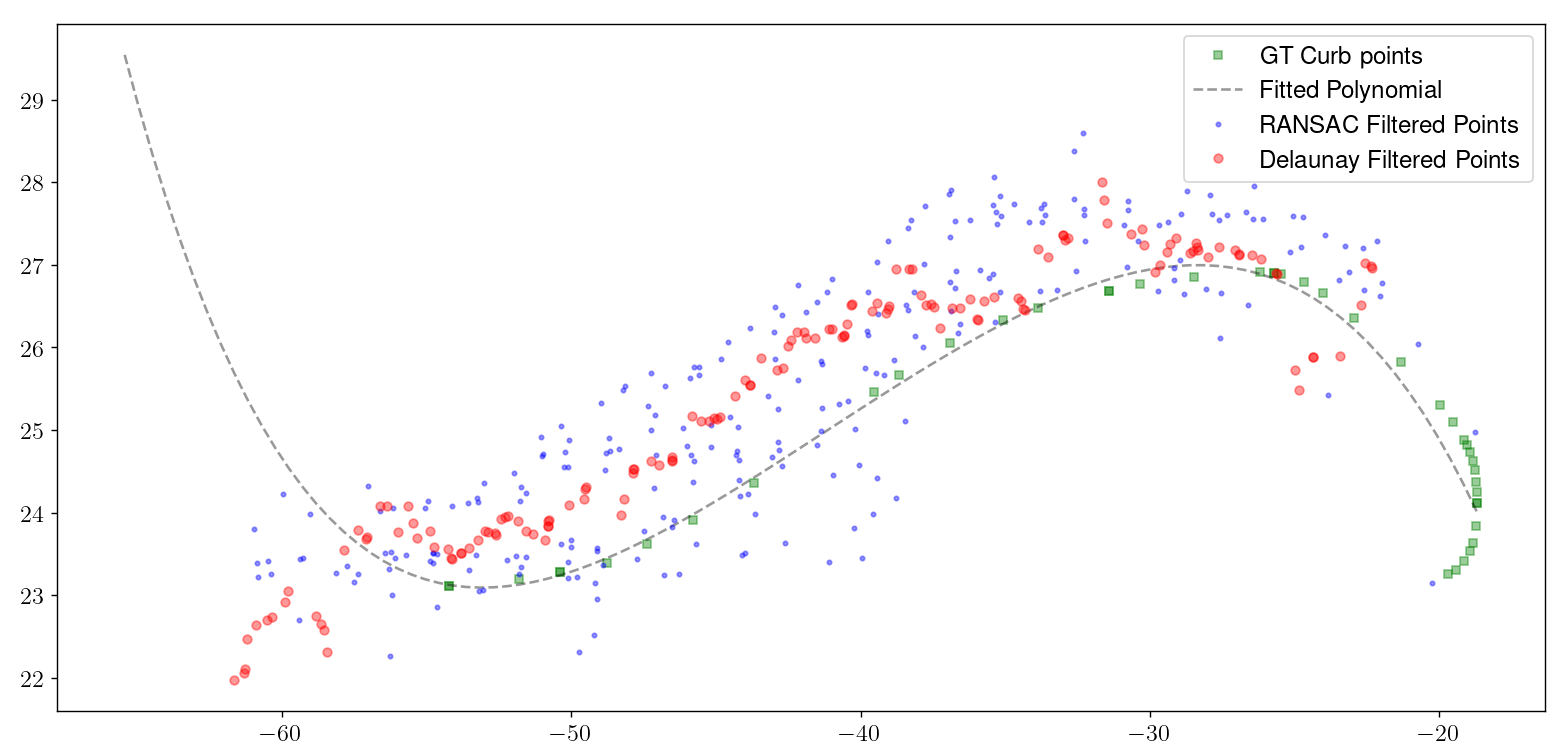}
		\caption{The curb points are extracted by applying RANSAC (blue points) and Delaunay filtering (red points). A polynomial fitting is performed (black line) to the GT points (green) after which the $L_2$-norm between the GT polyline and the filtered points are calculated.}
		\label{fig:curb_fitting}
		\vspace{-0.5cm}
	\end{figure}

	%\mfallon{again ... this is lack of ordering is not part of the science of this paper. Just because you/Scania havent parsed the curb database fully doesnt mean you can present this data clearning as a component of the work. Please remove this and just let the user assume the associations exist. Perhaps mention that its needed ... but not more than that}
	%\mfallon{that said, I see that you have limited experimental analysis to present so perhaps that is why you have included it}
	We associate the GT points from the map supplier segment-wise. For evaluation, we fit a polynomial to the GT points. Then we sample points from the polynomial and associate them to the Delaunay filtered points and RANSAC filtered points as shown in \autoref{fig:curb_fitting}. We compute the normalized $L_2$-norm (based on the number of points selected) for evaluation metrics. 

	\subsection{Automatic segment-wise association and evaluation}
	To automatically evaluate the generated curb points we measure the Chamfer distance (CD) \cite{barrow1977parametric} of each cluster segment. The CD between two set of point clouds $P_{1}, P_{2} \subseteq \mathbb{R}^{3}$ is defined as:
	\begin{equation}
	\begin{aligned}
	& \text{CD}\left(P_{1}, P_{2}\right) = \\ &\frac{1}{|P_1|}\sum_{a \in P_{1}} \min _{b \in P_{2}}\|a-b\|_{2}^{2} + \frac{1}{|P_2|}\sum_{b \in P_{2}} \min _{a \in P_{1}}\|b-a\|_{2}^{2}
	\end{aligned}
	\end{equation}
	We also evaluate the effect of different unsupervised clustering algorithms including Agglomerative Clustering \cite{beeferman2000agglomerative}, BIRCH \cite{zhang1996birch}, DBSCAN \cite{ester1996density} and OPTICS \cite{ankerst1999optics} for our evaluation metrics. The CD and the number of detected filtered curb points for our evaluation are reported in \autoref{tab:analysis}. 
	
	%Analysis of results
	We observe that Delaunay filtering identifies inliers that more closely correspond to the GT points than a generic RANSAC-based filtering both for manual and automatic segment wise association as the $L_2$-norm and CD are lower for Delaunay filtering compared to RANSAC. We also observe that the Delaunay filtering selects lesser inlier points compared to RANSAC. However, the inliers correspond to the GT points more closely as shown in \autoref{fig:curb_fitting}. We conclude that density based unsupervised algorithm for segment clustering works best for fitting an arbitrary number of curbs based on the computed CD. The result of our proposed method showing the final selected curb points is shown in \autoref{fig:curb_traj}.

	%\mfallon{IMPORTANT: you have no summary of the algorithmic performance!!! no intuition given about how well it worked and why it worked or what was suboptimal. There is no evaluation of the numbers at all!}
	%\mfallon{I cannot see a trend/pattern in what you did so I doubt the reader will be able to either}
	
	%https://stackoverflow.com/questions/64911820/fit-curve-spline-to-3d-point-cloud
	\begin{table}[!h]
	\vspace{-0.1cm}
	\centering
	\resizebox{\columnwidth}{!}{
	\begin{tabular}{l|cc}  \toprule
	\multicolumn{3}{c}{\textbf{Manual segment-wise association}}
	\\
	\midrule
	\textbf{No Clustering} & \textbf{Normalized $L_2$-Norm} & 
	\textbf{\# Detected Points}\\
	\midrule
	RANSAC Filtering & 27.659 & 9578\\
	Delaunay Filtering & \textbf{19.947} & 6904\\
%	\midrule
%	\textbf{Automatic Association} & \textbf{Chamfer Distance}& \textbf{\# Detected Points}\\
%	\midrule
%	No Clustering & 13.656 & 13904\\
%	Agglomerative Clustering& 13.656 & 13904\\
%	BIRCH & 13.656 & 13904\\
%	DBSCAN & 15.198 & 11312\\
%	OPTICS & 15.298 & 8280\\
	\midrule
	\multicolumn{3}{c}{\textbf{Automatic segment-wise association}} \\
	\midrule
	\textbf{Outlier Removal (RANSAC)} & \textbf{Chamfer Distance}& \textbf{\# Detected Points}\\
	\midrule
	%No Clustering & 179.194 & 33 \\
	Agglomerative Clustering & 17.427 & 3489 \\
	BIRCH & 19.596 & 1351\\
	DBSCAN & \textbf{17.220} & 5314\\
	OPTICS & 18.370 & 7446\\
	\midrule
	\textbf{Outlier Removal (Delaunay)} & \textbf{Chamfer Distance}& \textbf{\# Detected Points}\\
	\midrule
	Agglomerative Clustering & 15.418 & 3924 \\
	BIRCH & 16.165 & 3492\\
	DBSCAN & \textbf{14.753} & 6678\\
	OPTICS & 15.870 & 4415\\
	\bottomrule
	\end{tabular}}
	\caption{Evaluation Metrics}
	\label{tab:analysis}
	\vspace{-6mm}
	\end{table}
	\begin{comment}
	\begin{figure} [!htbp]
	\centering
	\includegraphics[width=0.49\columnwidth]{figures/images/raw-front.png}
	\includegraphics[width=0.49\columnwidth]{figures/images/seg-front.png} \\
	\vspace{0.013\columnwidth}
	\includegraphics[width=0.49\columnwidth]{figures/images/raw-back.png}
	\includegraphics[width=0.49\columnwidth]{figures/images/seg-back.png}
	\caption{Semantic segmentation results using modified EfficientNet-B0 network architecture as shown in Table \ref{tab:efficientNet}. \textit{Top:} Front camera raw image and its corresponding segmentation. \textit{Bottom:} Results from rear camera.}
	\label{fig:semseg-results}
	\vspace{-3mm}
	\end{figure}
	\end{comment}

	%medial axis: https://www.youtube.com/watch?v=tdVHKzYt5TM
	
	\section{Conclusion}
	\label{sec:conclusion}
	% contribution summary
	We proposed a multi-modal curb detection and mapping algorithm with a novel filtering approach using 3D-Delaunay tetrahedra. We demonstrated the detection of arbitrary number of curbs with our clustering approach. Our evaluation indicates that Delaunay filtering outperforms traditional RANSAC based filtering approach for curb outlier removal.

    %future work
	To extend this work further, we would study the instability of the medial axis generation under different noise conditions. We would also like to extend the semantic association to other infrastructure features like road lines, traffic lights, pedestrian paths. We also strive to benchmark our solution on open source datasets by retraining our semantic segmentation model on un-distorted images.  %We would also like to extend this framework towards a multi-session semantic mapping capability to create feature maps of an environment in global frame. This will ensure we could build HD-maps of an environment from multiple vehicles. 		

	\bibliographystyle{./IEEEtran}
	\bibliography{./IEEEabrv, ./library}

\begin{thebibliography}{10}
\providecommand{\url}[1]{#1}
\csname url@rmstyle\endcsname
\providecommand{\newblock}{\relax}
\providecommand{\bibinfo}[2]{#2}
\providecommand\BIBentrySTDinterwordspacing{\spaceskip=0pt\relax}
\providecommand\BIBentryALTinterwordstretchfactor{4}
\providecommand\BIBentryALTinterwordspacing{\spaceskip=\fontdimen2\font plus
\BIBentryALTinterwordstretchfactor\fontdimen3\font minus
  \fontdimen4\font\relax}
\providecommand\BIBforeignlanguage[2]{{%
\expandafter\ifx\csname l@#1\endcsname\relax
\typeout{** WARNING: IEEEtran.bst: No hyphenation pattern has been}%
\typeout{** loaded for the language `#1'. Using the pattern for}%
\typeout{** the default language instead.}%
\else
\language=\csname l@#1\endcsname
\fi
#2}}

\bibitem{BarHillel2014}
A.~Bar~Hillel, R.~Lerner, D.~Levi, and G.~Raz, ``Recent progress in road and
  lane detection: a survey,'' \emph{Machine Vision and Applications}, vol.~25,
  no.~3, pp. 727--745, Apr. 2014.

\bibitem{Wijesoma_2004}
W.~Wijesoma, K.~Kodagoda, and A.~Balasuriya, ``Road-boundary detection and
  tracking using ladar sensing,'' \emph{IEEE Transactions on Robotics and
  Automation}, vol.~20, no.~3, pp. 456--464, 2004.

\bibitem{Wang2019}
G.~Wang, J.~Wu, R.~He, and S.~Yang, ``A point cloud-based robust road curb
  detection and tracking method,'' \emph{IEEE Access}, vol.~7, pp.
  24\,611--24\,625, 2019.

\bibitem{fernandez2014road}
C.~Fern{\'a}ndez, R.~Izquierdo, D.~F. Llorca, and M.~Sotelo, ``Road curb and
  lanes detection for autonomous driving on urban scenarios,'' in \emph{17th
  International IEEE Conference on Intelligent Transportation Systems
  (ITSC)}.\hskip 1em plus 0.5em minus 0.4em\relax IEEE, 2014, pp. 1964--1969.

\bibitem{chen2015velodyne}
T.~Chen, B.~Dai, D.~Liu, J.~Song, and Z.~Liu, ``Velodyne-based curb detection
  up to 50 meters away,'' in \emph{2015 IEEE Intelligent Vehicles Symposium
  (IV)}.\hskip 1em plus 0.5em minus 0.4em\relax IEEE, 2015, pp. 241--248.

\bibitem{oniga2010polynomial}
F.~Oniga and S.~Nedevschi, ``Polynomial curb detection based on dense
  stereovision for driving assistance,'' in \emph{13th International IEEE
  Conference on Intelligent Transportation Systems}.\hskip 1em plus 0.5em minus
  0.4em\relax IEEE, 2010, pp. 1110--1115.

\bibitem{siegemund2010curb}
J.~Siegemund, D.~Pfeiffer, U.~Franke, and W.~F{\"o}rstner, ``Curb
  reconstruction using conditional random fields,'' in \emph{2010 IEEE
  Intelligent Vehicles Symposium}.\hskip 1em plus 0.5em minus 0.4em\relax IEEE,
  2010, pp. 203--210.

\bibitem{kellner2014road}
M.~Kellner, M.~E. Bouzouraa, and U.~Hofmann, ``Road curb detection based on
  different elevation mapping techniques,'' in \emph{2014 IEEE Intelligent
  Vehicles Symposium Proceedings}.\hskip 1em plus 0.5em minus 0.4em\relax IEEE,
  2014, pp. 1217--1224.

\bibitem{goga2018fusing}
S.~E.~C. Goga and S.~Nedevschi, ``Fusing semantic labeled camera images and 3d
  lidar data for the detection of urban curbs,'' in \emph{2018 IEEE 14th
  International Conference on Intelligent Computer Communication and Processing
  (ICCP)}.\hskip 1em plus 0.5em minus 0.4em\relax IEEE, 2018, pp. 301--308.

\bibitem{deac2019fusion}
S.~E. Catalina~Deac, I.~Giosan, and S.~Nedevschi, ``Curb detection in urban
  traffic scenarios using lidars point cloud and semantically segmented color
  images,'' in \emph{2019 IEEE Intelligent Transportation Systems Conference
  (ITSC)}, 2019, pp. 3433--3440.

\bibitem{baek2020curbscan}
I.~Baek, T.-C. Tai, M.~M. Bhat, K.~Ellango, T.~Shah, K.~Fuseini, and R.~R.
  Rajkumar, ``Curbscan: Curb detection and tracking using multi-sensor
  fusion,'' in \emph{2020 IEEE 23rd International Conference on Intelligent
  Transportation Systems (ITSC)}.\hskip 1em plus 0.5em minus 0.4em\relax IEEE,
  2020, pp. 1--8.

\bibitem{aufrere2003multiple}
R.~Aufrere, C.~Mertz, and C.~Thorpe, ``Multiple sensor fusion for detecting
  location of curbs, walls, and barriers,'' in \emph{IEEE IV2003 Intelligent
  Vehicles Symposium. Proceedings (Cat. No. 03TH8683)}.\hskip 1em plus 0.5em
  minus 0.4em\relax IEEE, 2003, pp. 126--131.

\bibitem{kim2007outdoor}
S.-H. Kim, C.-W. Roh, S.-C. Kang, and M.-Y. Park, ``Outdoor navigation of a
  mobile robot using differential gps and curb detection,'' in
  \emph{Proceedings 2007 IEEE International Conference on Robotics and
  Automation}.\hskip 1em plus 0.5em minus 0.4em\relax IEEE, 2007, pp.
  3414--3419.

\bibitem{stuckler2008lane}
J.~Stuckler, H.~Schulz, and S.~Behnke, ``In-lane localization in road networks
  using curbs detected in omnidirectional height images,'' \emph{VDIBERICHT},
  vol. 2012, p. 151, 2008.

\bibitem{el2011detection}
S.~El-Halawany, A.~Moussa, D.~D. Lichti, and N.~El-Sheimy, ``Detection of road
  curb from mobile terrestrial laser scanner point cloud,'' \emph{International
  Archives of the Photogrammetry, Remote Sensing and Spatial Information
  Sciences}, vol.~38, no. 5/W12, 2011.

\bibitem{wisth2021unified}
D.~Wisth, M.~Camurri, S.~Das, and M.~Fallon, ``Unified multi-modal landmark
  tracking for tightly coupled lidar-visual-inertial odometry,'' \emph{IEEE
  Robotics and Automation Letters}, vol.~6, no.~2, pp. 1004--1011, 2021.

\bibitem{alexnet_2012}
A.~Krizhevsky, I.~Sutskever, and G.~E. Hinton, ``Imagenet classification with
  deep convolutional neural networks,'' \emph{NIPS}, vol.~25, 2012.

\bibitem{tan2019efficientnet}
M.~Tan and Q.~Le, ``Efficientnet: Rethinking model scaling for convolutional
  neural networks,'' in \emph{International Conference on Machine
  Learning}.\hskip 1em plus 0.5em minus 0.4em\relax PMLR, 2019, pp. 6105--6114.

\bibitem{sandler2018mobilenetv2}
M.~Sandler, A.~Howard, M.~Zhu, A.~Zhmoginov, and L.-C. Chen, ``Mobilenetv2:
  Inverted residuals and linear bottlenecks,'' in \emph{Proceedings of the IEEE
  conference on computer vision and pattern recognition}, 2018, pp. 4510--4520.

\bibitem{hu2018squeeze}
J.~Hu, L.~Shen, and G.~Sun, ``Squeeze-and-excitation networks,'' in \emph{IEEE
  Conference on Computer Vision and Pattern Recognition (CVPR)}, 2018, pp.
  7132--7141.

\bibitem{kannala2006generic}
J.~Kannala and S.~S. Brandt, ``A generic camera model and calibration method
  for conventional, wide-angle, and fish-eye lenses,'' \emph{IEEE transactions
  on pattern analysis and machine intelligence}, vol.~28, no.~8, pp.
  1335--1340, 2006.

\bibitem{ester1996density}
M.~Ester, H.-P. Kriegel, J.~Sander, X.~Xu, \emph{et~al.}, ``A density-based
  algorithm for discovering clusters in large spatial databases with noise.''
  in \emph{kdd}, vol.~96, no.~34, 1996, pp. 226--231.

\bibitem{brandt1992continuous}
J.~W. Brandt and V.~R. Algazi, ``Continuous skeleton computation by voronoi
  diagram,'' \emph{CVGIP: Image understanding}, vol.~55, no.~3, pp. 329--338,
  1992.

\bibitem{vornoi_mp_1989}
O.~Takahashi and R.~Schilling, ``Motion planning in a plane using generalized
  voronoi diagrams,'' \emph{IEEE Transactions on Robotics and Automation},
  vol.~5, no.~2, pp. 143--150, 1989.

\bibitem{garber2004constraint}
M.~Garber and M.~C. Lin, ``Constraint-based motion planning using voronoi
  diagrams,'' in \emph{Algorithmic Foundations of Robotics V}.\hskip 1em plus
  0.5em minus 0.4em\relax Springer, 2004, pp. 541--558.

\bibitem{fortune1995voronoi}
S.~Fortune, ``Voronoi diagrams and delaunay triangulations,'' \emph{Computing
  in Euclidean geometry}, pp. 225--265, 1995.

\bibitem{ransac_1981}
M.~A. Fischler and R.~C. Bolles, ``Random sample consensus: A paradigm for
  model fitting with applications to image analysis and automated
  cartography,'' \emph{Commun. ACM}, vol.~24, no.~6, p. 381–395, jun 1981.

\bibitem{zhou2018open3d}
Q.-Y. Zhou, J.~Park, and V.~Koltun, ``Open3d: A modern library for 3d data
  processing,'' \emph{arXiv preprint arXiv:1801.09847}, 2018.

\bibitem{scikit-learn}
F.~Pedregosa, G.~Varoquaux, A.~Gramfort, V.~Michel, B.~Thirion, O.~Grisel,
  M.~Blondel, P.~Prettenhofer, R.~Weiss, V.~Dubourg, J.~Vanderplas, A.~Passos,
  D.~Cournapeau, M.~Brucher, M.~Perrot, and E.~Duchesnay, ``Scikit-learn:
  Machine learning in {P}ython,'' \emph{Journal of Machine Learning Research},
  vol.~12, pp. 2825--2830, 2011.

\bibitem{barrow1977parametric}
H.~G. Barrow, J.~M. Tenenbaum, R.~C. Bolles, and H.~C. Wolf, ``Parametric
  correspondence and chamfer matching: Two new techniques for image matching,''
  Tech. Rep., 1977.

\bibitem{beeferman2000agglomerative}
D.~Beeferman and A.~Berger, ``Agglomerative clustering of a search engine query
  log,'' in \emph{Proceedings of the sixth ACM SIGKDD international conference
  on Knowledge discovery and data mining}, 2000, pp. 407--416.

\bibitem{zhang1996birch}
T.~Zhang, R.~Ramakrishnan, and M.~Livny, ``Birch: an efficient data clustering
  method for very large databases,'' \emph{ACM sigmod record}, vol.~25, no.~2,
  pp. 103--114, 1996.

\bibitem{ankerst1999optics}
M.~Ankerst, M.~M. Breunig, H.-P. Kriegel, and J.~Sander, ``Optics: Ordering
  points to identify the clustering structure,'' \emph{ACM Sigmod record},
  vol.~28, no.~2, pp. 49--60, 1999.

\end{thebibliography}
	
	%%Bibtex cleaner: https://flamingtempura.github.io/bibtex-tidy/
	
\end{document}